%% file: main.tex
\documentclass[sigconf,nonacm,authorversion]{acmart}

\usepackage{multirow} 
\usepackage{natbib}
\usepackage{pifont}
\usepackage{bm}
\usepackage{textcomp} 
\usepackage{multirow}
\usepackage{colortbl}
\usepackage{colortbl}
\usepackage{amsmath}
\usepackage{mathrsfs}
\usepackage{amsmath} 
\usepackage{color}
\usepackage{adjustbox}
\usepackage{url}
\usepackage{bbding}
 
\usepackage{amssymb}
\usepackage{bm}
\usepackage{graphicx}
\newcommand{\myparagraph}[1]{\textbf{#1}\hspace{1.8ex}}
\definecolor{aliceblue}{rgb}{0.87, 0.92, 0.96}
\newcommand{\largemodel}[1]{\color{gray}{#1}}

\begin{document}

\title{MediSee: Reasoning-based Pixel-level Perception in Medical Images}

\author{Qinyue Tong}
\email{22224030@zju.edu.cn}

\affiliation{%
  \institution{Zhejiang University}
  \city{Hangzhou}
  \state{Zhejiang}
  \country{China}
}

\author{Ziqian Lu}
\authornote{corresponding author}
\email{ziqianlu@zju.edu.cn}

\affiliation{%
  \institution{Zhejiang Sci-Tech University}
  \city{Hangzhou}
  \state{Zhejiang}
  \country{China}
}

\author{Jun Liu}
\email{22224051@zju.edu.cn}

\affiliation{%
  \institution{Zhejiang University}
  \city{Hangzhou}
  \state{Zhejiang}
  \country{China}
}

\author{Yangming Zheng}
\email{zymsun2002@zju.edu.cn}

\affiliation{%
  \institution{Zhejiang University}
  \city{Hangzhou}
  \state{Zhejiang}
  \country{China}
}

\author{Zhe-ming Lu}
\email{zheminglu@zju.edu.cn}

\affiliation{%
  \institution{Zhejiang University}
  \city{Hangzhou}
  \state{Zhejiang}
  \country{China}
}

\renewcommand{\shortauthors}{Qinyue Tong et al.}

\begin{abstract}

Despite remarkable advancements in pixel-level medical image perception, existing methods are either limited to specific tasks or heavily rely on accurate bounding boxes or text labels as input prompts.
However, the medical knowledge required for input is a huge obstacle for general public, which greatly reduces the universality of these methods. Compared with these domain-specialized auxiliary information, general users tend to rely on oral queries that require logical reasoning.
In this paper, we introduce a novel medical vision task: Medical Reasoning Segmentation and Detection (\textbf{MedSD}), which aims to comprehend implicit queries about medical images and generate the corresponding segmentation mask and bounding box for the target object.
To accomplish this task, we first introduce a Multi-perspective, Logic-driven Medical Reasoning Segmentation and Detection (\textbf{MLMR-SD}) dataset, which encompasses a substantial collection of medical entity targets along with their corresponding reasoning.
Furthermore, we propose \textbf{MediSee}, an effective baseline model designed for medical reasoning segmentation and detection. 
The experimental results indicate that the proposed method can effectively address MedSD with implicit colloquial queries and outperform traditional medical referring segmentation methods.
Code is available at \textit{\textcolor{cyan}{\url{https://github.com/Edisonhimself/MediSee}}}.

\end{abstract}



\begin{teaserfigure}
  \centering
  \includegraphics[width=0.87\textwidth]{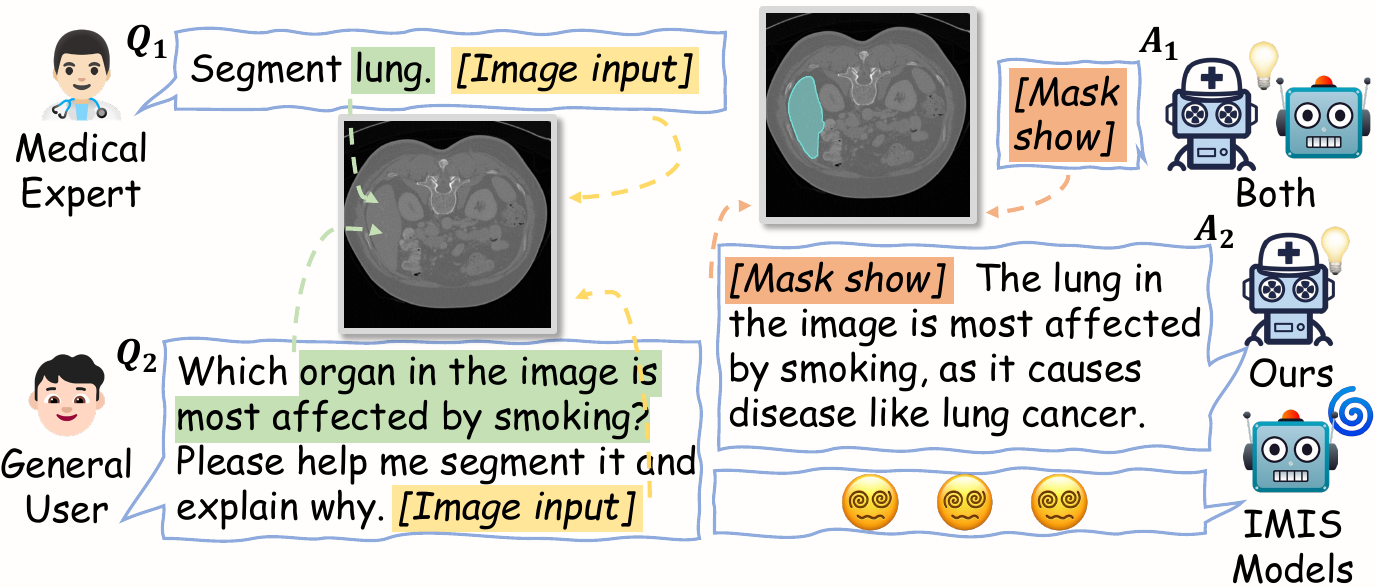}
  \vspace{-0.9em}
  \caption{The demo dialogue of our proposed MediSee. Our model is capable of handling both referring expressions and queries that require complex reasoning and domain-specific medical knowledge. 
   In contrast, vanilla IMIS models are restricted to addressing only referring instructions and are unable to infer implicit queries related to medical images.}
  \Description{The demo dialogue of our proposed MediSee. Our model is capable of handling both referring expressions and queries that require complex reasoning and domain-specific medical knowledge,
while vanilla IMIS models are limited to addressing only the former.}
  \label{fig:teaser}
\end{teaserfigure}

\maketitle

\begin{figure*}
    \centering
    \includegraphics[width=0.91\linewidth]{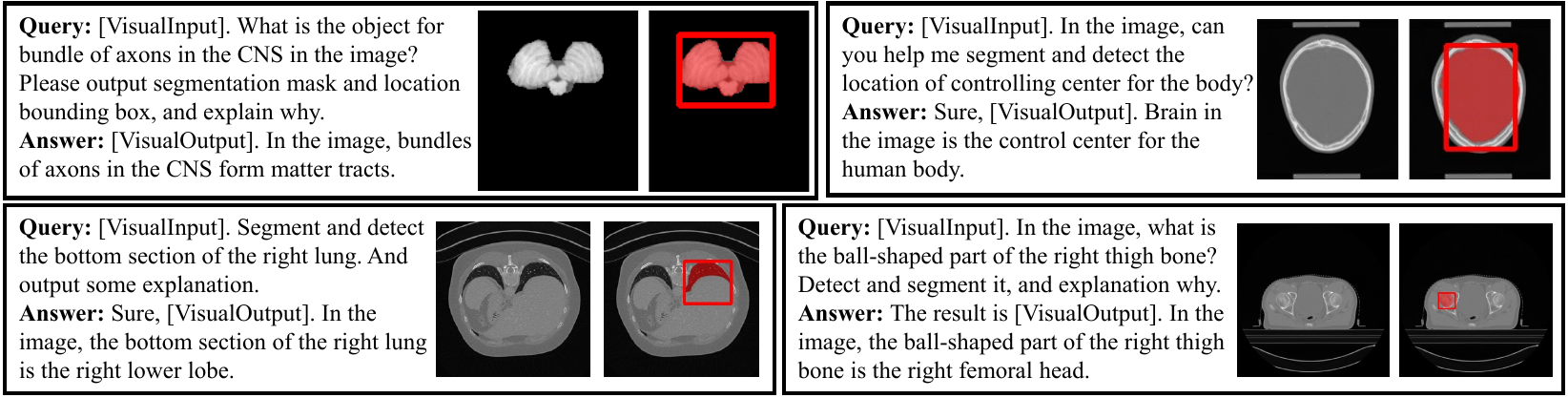}
    \vspace{-1.1em}
    \caption{We introduce MediSee, a model that opens the door to medical reasoning image perception, capable of handling cases that demand complex reasoning and domain-specific medical knowledge. Notably, our model is also able to generate corresponding textual explanations for the given queries, enhancing the interpretability and interactivity.}
    \vspace{-0.5em}
    \label{fig:visual_perform_intro}
    \Description{}
\end{figure*}

\section{Introduction}
Medical image segmentation seeks to identify regions of interest (ROIs), such as organs, lesions, and tissue, within a variety of medical images \cite{instruction-med-seg-onesentence-1,intro_first_seg-2,intro_first_seg-3}. This process is pivotal in numerous clinical applications, including disease diagnosis, treatment planning, and monitoring disease progression \cite{intro-disease-progression-tracking-1,intro-disease-progression-tracking-2,intro-disease-progression-tracking-3,intro-disease-progression-tracking-4}, as well as in advancing medical research \cite{intro-medical-researches-1}.
Previous works predominantly focus on task-specific segmentation \cite{intro-specialist-model-1,intro-specialist-model-2,related-works-unet}, with these methods commonly referred to as ``specialist models'' . 
Although demonstrating impressive segmentation performance in specific tasks, such approaches remain constrained by insufficient flexibility and a lack of interactive capabilities in real-world applications.

Recent advancements in Interactive Medical Image Segmentation (IMIS) overcome this limitation by allowing users to actively guide the model via intuitive annotations, such as points, texts, or regions, thereby enabling more diverse and user-driven segmentation outcomes \cite{intro-imis-1,intro-imis-2}.
However, the usability of these methods remains a significant challenge, as many state-of-the-art IMIS tools require users to provide precise bounding boxes or labeled texts to define regions of interest \cite{One-model-to-rule-them-all,medsam_model,sam-med-2d-model,biomed-parse,imis_model}, thereby placing a considerable demand on users' medical expertise.
In contrast, general users often rely on implicit queries, requiring logical reasoning just like the process of Chain of Thought (CoT) \cite{chain-of-thought}. 
As illustrated in Figure \ref{fig:teaser}, compared to professional medical input (\emph{e.g., the class name of lung}) \(Q_{1}\), general users pay more attention on throwing an implicit query (\emph{e.g.,} $Q_2$). 
Unfortunately, due to the lack of medical reasoning abilities, existing IMIS methods usually fail to give expected responses for these inputs that involve implicit logic. 


In this work, we propose a novel medical vision task---Medical Reasoning Segmentation and Detection (\textbf{MedSD})---which entails generating binary segmentation masks and bounding boxes based on implicit textual queries about medical images that involves complex reasoning. 
Notably, the query is not restricted to a direct reference but may encompass more intricate descriptions requiring advanced reasoning or domain-specific medical knowledge.
To successfully perform this task, the model should equip two essential capabilities: (1) Comprehending and reasoning over complex and implicit medical text queries in conjunction with the images. (2) Generating accurate segmentation masks and bounding boxes while providing relevant explanations in response to the query.

Given that data scarcity represents a major bottleneck in accomplishing MedSD, we first introduce a Multi-perspective, Logic-driven Medical Reasoning Segmentation and Detection dataset termed \textbf{MLMR-SD}, which includes a substantial number of medical entity targets along with their associated reasoning.
We collect images, masks, and label annotations from the publicly available SA-Med2D-20M \cite{sam_med_2d_20m}.
Subsequently, drawing inspiration from LLaVA-Med \cite{llava-med}, we generate intricate question-answer pairs using GPT-4 \cite{gpt-4} and LLaVA-Med. 
As a result, the MLMR-SD dataset comprises a vast and diverse set of 200K complex and implicit instructions for medical reasoning segmentation and detection.
Notably, a key distinction between the proposed MLMR-SD and traditional medical segmentation datasets \cite{intro-traditional-seg-dataset-1,intro-traditional-seg-dataset-2,intro-traditional-seg-dataset-3,traditional-seg-dataset-1,traditional-seg-dataset-2} is that our dataset includes rich, meaningful, logic-driven question-answer pairs generated from images and labels. 
This not only enhances the practical applicability of the MLMR-SD but also promotes greater interactivity in the segmentation masks and bounding boxes generating.

\begin{figure*}
    \centering
    \includegraphics[width=1\linewidth]{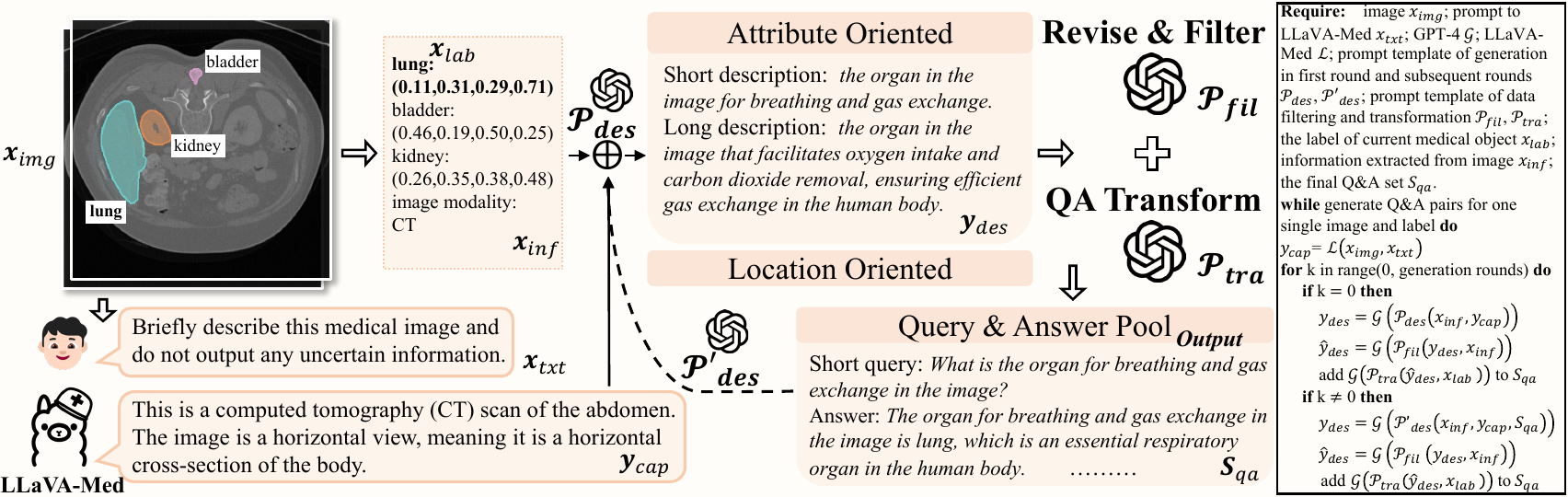}
    \vspace{-2.1em}
    \caption{The data generation pipeline of MLMR-SD dataset (left) and its pseudocode (right). The figure illustrates the process of generating question-answer pairs for the medical object ``lung'' in the image. 
    All variables are defined in the pseudocode.
    }
    \vspace{-0.5em}
    \label{fig:dataset_pipeline}
    \Description{}
\end{figure*}

\begin{figure}
    \centering
    \includegraphics[width=1\linewidth]{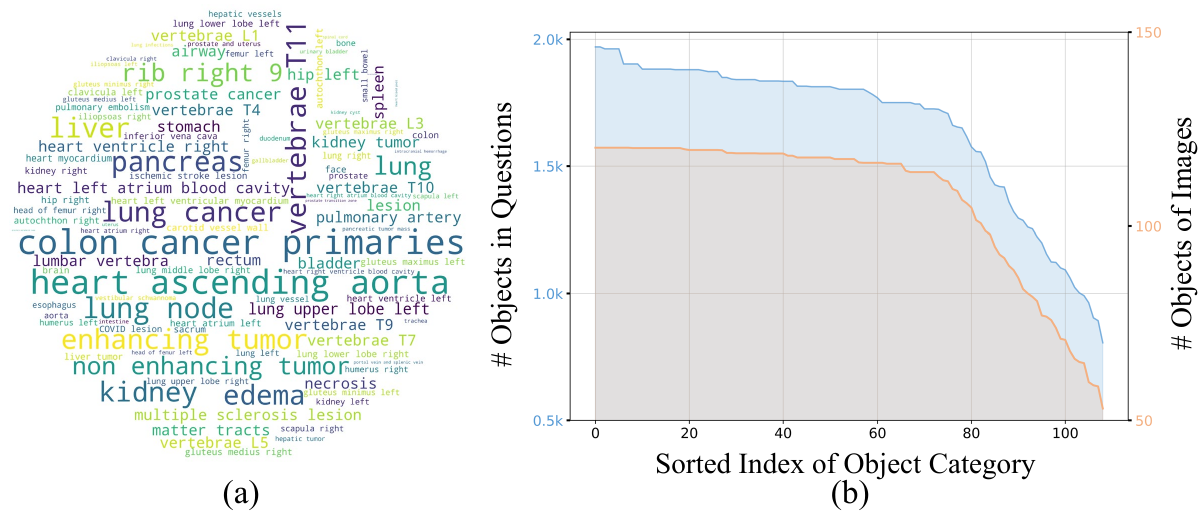}
    \vspace{-2.1em}
    \caption{Analysis of the data structure in MLMR-SD. (a) word cloud in MLMR-SD; (b) the frequency distribution for each medical object across the question-answer pairs and images.}
    \vspace{-1.25em}
    \label{fig:dataset_structure}
    \Description{}
\end{figure}

Furthermore, building upon reasoning-based segmentation in natural image works \cite{intro-seg-llm,lisa,lisa++}, we introduce \textbf{MediSee}, an effective baseline model designed for medical reasoning segmentation and detection.
MediSee utilizes a democratic and adaptive feature fusion mechanism, leveraging the hidden representations of multiple candidate tokens to enhance the quality and diversity of the candidate guidance information input to the downstream decoder.
This strategy enables the model to more effectively capture the correspondence between medical entities in implicit queries and their spatial locations within the images, while balancing the collaborative performance across downstream decoders.
Moreover, during the additional fine-tuning on MediSee, we incorporate a novel similarity map-based supervision signal, further enhancing the model's performance on the MLMR-SD dataset.
As shown in Figure \ref{fig:visual_perform_intro}, MediSee can reason complex queries about medical images.
Experimental results on different benchmarks, including MLMR-SD and traditional medical referring segmentation dataset, demonstrate the superiority of our model.
We believe that our dataset and model constitute a valuable resource for real-world medical reasoning image perception, providing enhanced versatility and robustness for a wide range of applications.

In summary, our contributions are as follows:
\begin{itemize}
    \item We introduce the medical reasoning segmentation and detection task (MedSD), which requires reasoning output based on user queries about medical images. 
    \item We construct the MLMR-SD dataset, which includes 200K complex and implicit question pairs for MedSD. This dataset serves as a strong foundation for research on the MedSD.
    \item We propose an effective baseline model, MediSee to adaptively fuse multiple candidate guidance information. Extensive experiments demonstrate the outstanding performance of our model, both in MedSD, as well as in traditional medical referring segmentation and detection.
\end{itemize}

\section{Related Works}
\noindent \myparagraph{Medical Image Segmentation.}
The U-Net series \cite{related-works-unet,related-works-unet++,related-works-ResU-Net,related-works-ResU-nnU-Net} represents one of the most classical architectures for medical image segmentation. However, these models are typically task-specific and fully automated, exhibiting strong performance on particular imaging modalities or targets, yet often failing to generalize across diverse scenarios. 
Therefore, their limited flexibility and user interactivity impose substantial limitations on their adaptability and practical application in real-world settings.
The development of Interactive Medical Image Segmentation (IMIS) models alleviate this limitation by enabling clinicians or users to guide the segmentation process through specific forms of input \cite{intro-imis-1,intro-imis-2}, such as scribbles \cite{related-works-Scribble-1,related-works-Scribble-2}, bounding boxes \cite{related-works-bbox}, clicks \cite{intro-imis-2,related-works-cliks-2}, or language prompts \cite{segvol,related-works-text-1}.
Although these methods demonstrate significant advantages in traditional IMIS, they lack the ability to handle implicit, reasoning-based textual queries, which poses considerable challenges for users without medical expertise.
In this work, we focus on reasoning-based medical image perception (segmentation and detection), with the goal of enabling model to comprehend implicit user queries while simultaneously generating accurate masks and bounding boxes for the medical target objects.

\noindent \myparagraph{Reasoning Segmentation in Natural Images.}
The task of reasoning segmentation in natural images is introduced by LISA \cite{lisa}, which aims to understand implicit text instructions and provide a corresponding mask for the given query. 
LISA preliminarily achieves the interpretation of implicit segmentation queries by combining exceptional visual-linguistic understanding of MLLMs \cite{llava} with the segmentation abilities of SAM \cite{segment-anything}.
Recently, several follow-up works on reasoning segmentation have emerged, including those addressing multi-scale objects \cite{mmr-iclr2025}, multi-round dialogue \cite{multi-round-seg}, and multi-object segmentation \cite{glamm-cvpr2024}, among others \cite{perceptiongpt-cvpr2024,next-chat}.
Compared to reasoning segmentation in natural domains, reasoning segmentation in medical imaging may hold greater practical value. 
This is because general users often possess common sense but lack specialized medical knowledge, which makes them more reliant on implicit reasoning-based queries when using medical segmentation models.
However, existing medical segmentation models are almost incapable of understanding implicit queries. In this paper, building upon existing works \cite{llava-med,medsam_model,sam_med_2d_20m,lisa}, we implement a method for perceiving medical images through reasoning queries, thereby enhancing the human-machine interaction and practical value of medical segmentation models.

\begin{figure*}
    \centering
    \includegraphics[width=0.93\linewidth]{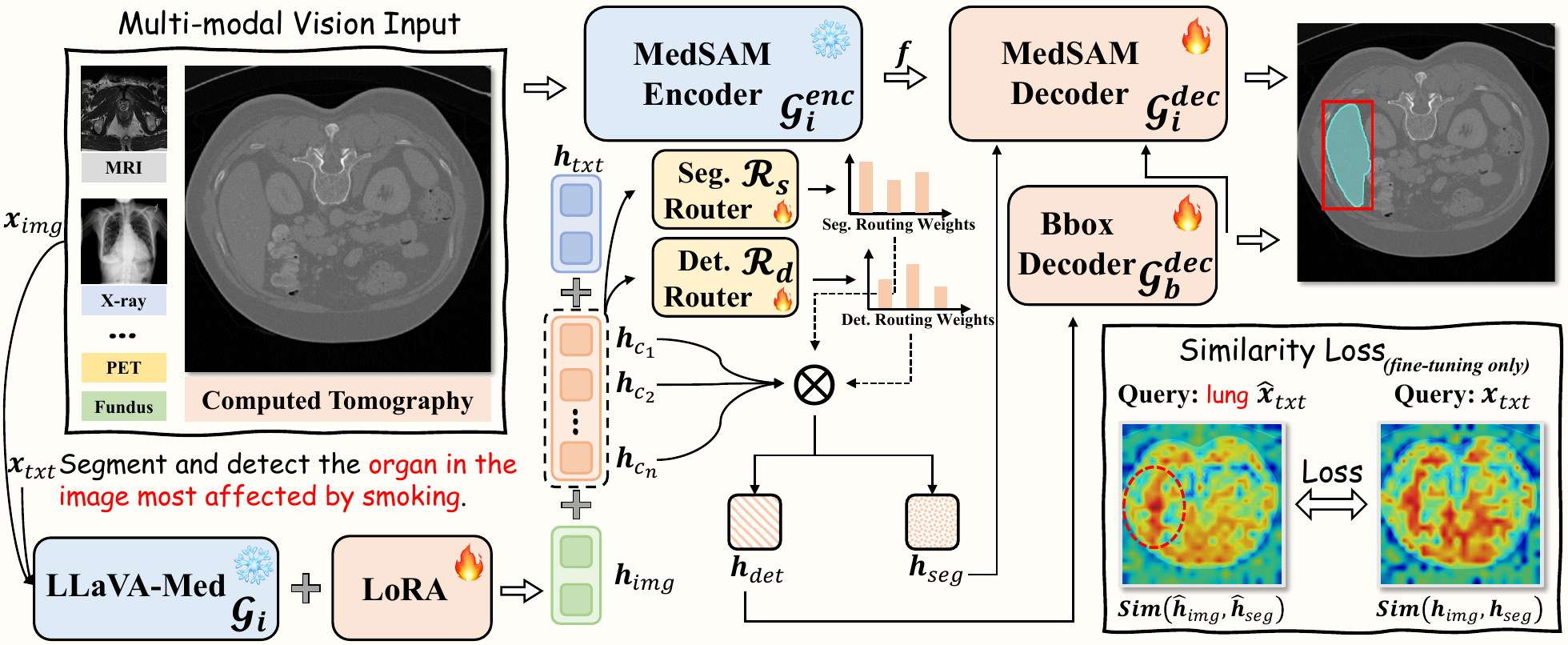}
    \vspace{-0.5em}
    \caption{The overview of MediSee framework and the similarity loss introduced during the additional fine-tuning phase (the lower right corner). \(h_{c_i}\), \(h_{img}\) and \(h_{txt}\) indicate the candidate token embedding, image embedding and text embedding, which are all derived from the last hidden-layer of LLaVA-Med's output. \(Sim(\cdot)\) represents the computation of the dot product similarity. \(\hat{h}_{img}\) and \(\hat{h}_{seg}\) represent the image embedding and the input features for the mask decoder, which are obtained based on non-inference query \(\hat{x}_{txt}\).} 
    \vspace{-0.9em}
    \label{fig:model_main}
    \Description{}
\end{figure*}

\section{MLMR-SD Dataset} \label{sec:data-generation}
Current publicly available datasets for multi-modal medical segmentation primarily focus on direct human-computer interaction, including clicks, bounding boxes, and scribbles paradigms.
Consequently, existing medical segmentation models often struggle with text-based queries involving complex logic and advanced reasoning.
To address these limitations, we develop a Multi-perspective, Logic-driven Medical Reasoning Segmentation and Detection dataset termed MLMR-SD.
MLMR-SD is a comprehensive image-text dataset that includes images, masks and bounding box information derived from the publicly available SA-Med2D-20M dataset \cite{sam_med_2d_20m}, equipped with implicit question-answer pairs generated by the GPT-4 \cite{gpt-4}.
Compared to traditional medical segmentation datasets, MLMR-SD incorporates a large-scale question-answer pairs that consider complex reasoning from different perspectives.
We introduce our automated annotation pipeline of MLMR-SD and provide a statistical analysis in the following sections.

\noindent \myparagraph{Data Generation.}
We employ a subset of high-quality images, masks, and labels from SA-Med2D-20M \cite{sam_med_2d_20m} as the foundational data for MLMR-SD, thereby significantly reducing the effort of data pre-processing and annotation.
As shown on the left side of Figure \ref{fig:dataset_pipeline}, to enhance GPT-4's understanding of the image information and facilitate the generation of high-quality reasoning question-answer pairs, we first employ LLaVA-Med \cite{llava-med} to provide a brief description of the medical image.
Specifically, given an image \(x_{img}\) with mask information and label annotations, we ask LLaVA-Med to describe the image briefly.
 Subsequently, we input the combination of the caption \(y_{cap}\) generated from LLaVA-Med and image information \(x_{inf}\) (including label annotations, bounding boxes derived from the mask, and the corresponding image modality) into GPT-4 with a well-crafted prompt template \(\mathcal{P}_{des}\) as if GPT-4 could see the image (even though it only has access to the text).
To effectively mitigate the impact of hallucinations generated by LLaVA-Med, we implement two measures. First, we constrain LLaVA-Med to produce only simple descriptions and avoid uncertain information. Second, we instruct GPT-4 to prioritize label information over caption content when interpreting the image. In cases of discrepancy between the two, the label data is regarded as authoritative.

For one medical object (\emph{e.g., lung}), GPT-4 is instructed to generate data from two perspectives, producing both long and short logically guided synonymous expressions for each perspective to enhance the dataset's versatility and applicability: 
\textbf{1) Attribute Oriented:} Generate reasoning-based descriptions directed towards the target medical object by using its unique attributes, which distinguish it from other objects in the image. 
\textbf{2) Location Oriented:} Generate logic-driven descriptions by leveraging the object's relative positional relationship within the image.
For clearer presentation, we provide dataset samples in the \textit{supplementary materials}.

\noindent \myparagraph{Revising, Filtering, and Transformation.}
Despite the employment of meticulously crafted prompt for GPT-4, occasional deviations from established rules may lead to the generation of suboptimal descriptions. 
These deviations include overly direct and simplistic descriptions or those lacking a clear logical direction.
Consequently, we once again borrow the advanced comprehension capabilities of GPT-4, assigning it the role of an evaluator to revise and filter out suboptimal descriptions.
Notably, following the construction of the dataset, we conduct an additional round of manual screening to ensure the validity and rationality of the data.

We ask GPT-4 to transform the post-processed descriptions into questions and, by using the labels as the correct answers, generate explanatory responses, thereby producing question-answer pairs.
Subsequently, we employ a new prompt template \(\mathcal{P'}_{des}\) to guide GPT-4 in generating question-answer pairs that differ from the existing outputs.
The pseudocode for the generation pipeline of MLMR-SD is presented on the right side of Figure \ref{fig:dataset_pipeline}.

\noindent \myparagraph{Data Statistics.}
The MLMR-SD dataset comprises over 200K implicit and complex question-answer pairs that require logical reasoning, along with 12,652 image-mask pairs. 
For each image, we select one target object and ensure that there are no duplicate objects.
We allocate 10,129 image-mask pairs to the training set, 1,257 to the validation set, and 1,266 to the test set.

Additionally, the MLMR-SD dataset encompasses a comprehensive range of 109 medical object categories, including various types of organs, lesions, and tissue, among others.
The word cloud presented in Figure \ref{fig:dataset_structure} visually represents the distribution of labels within our dataset, highlighting the diverse range of medical entity categories upon which our question-answer pairs are based. 
 We also perform a thorough analysis of the frequency distribution for each medical object across the question-answer and image-mask pairs, as depicted in Figure \ref{fig:dataset_structure}. The result demonstrates the data diversity of the MLMR-SD, ensuring that both the questions and images are not biased toward specific categories.


\section{Baseline Framework}
We propose a novel and effective baseline architecture for medical reasoning segmentation and detection, called MediSee.
MediSee is capable of reasoning over implicit and complex queries in conjunction with medical images to achieve pixel-level perception.
We provide a detailed introduction of our proposed model in the following sections.


\subsection{Model Architecture} \label{sec:model_architecture}
Figure \ref{fig:model_main} presents the overall architecture of our MediSee, which consists of four core components: MedSAM \cite{medsam_model} as vision backbone \(\mathcal{G}_i^{enc}\) and mask decoder \(\mathcal{G}_i^{dec}\), LLaVA-Med \cite{llava-med} as multi-modal large language foundation model  \(\mathcal{G}_i\), two routers \(\mathcal{R}_s\) and \(\mathcal{R}_d\) consist of 2-layer MLPs, and a 3-layer MLP as bounding box decoder \(\mathcal{G}_b^{dec}\).

\begin{table*}[t]
    \centering
    \caption{Comparison of medical reasoning segmentation performance between MediSee and prior related methods. \(\mathcal{E}\) and \(\mathcal{B}\) denote the explanation output and bounding box decoder, respectively. ``ft'' indicates the further fine-tuning on MLMR-SD.} 
    \vspace{-0.2cm}
    \label{table:reason_seg}   
    \tabcolsep=0.23cm
    {
        \begin{tabular}{ l | c c c | c c c |c c c | c  }
            \toprule[1.2pt]
            
            \multirow{3}*{\textbf{Methods}} & \multicolumn{3}{c|}{\textbf{val}} & \multicolumn{6}{c|}{\textbf{test}}  & \multirow{3}*{\(\mathcal{E}\)/\(\mathcal{B}\)}     \\

            \specialrule{0em}{0pt}{1pt}
            \cline{2-10}
            \specialrule{0em}{0pt}{1pt}
            
            
            ~ & \multicolumn{3}{c|}{\textbf{overall}} & \multicolumn{3}{c|}{\textbf{short query}} & \multicolumn{3}{c|}{\textbf{long query}} & ~ \\

            \specialrule{0em}{0pt}{1pt}
            \cline{2-10}
            \specialrule{0em}{0pt}{1pt}
            
            ~ & \textbf{Dice} & \textbf{gIoU} & \textbf{cIoU} & \textbf{Dice} & \textbf{gIoU} & \textbf{cIoU} & \textbf{Dice} & \textbf{gIoU} & \textbf{cIoU} & ~  \\ 
            
            \specialrule{0em}{0pt}{1pt}
            \hline
            \hline
            \specialrule{0em}{0pt}{1pt}

            Grounding DINO \cite{grounding-dino} + SAM-Med2D \cite{sam-med-2d-model} & 15.1 & 9.9 & 8.4 &10.3  &9.0  &10.2  &10.6  &8.5  &10.1   & \multirow{4}*{\XSolidBrush/\XSolidBrush}\\

            Grounding DINO \cite{grounding-dino} + MedSAM \cite{medsam_model} &12.9  &8.7  &12.4  &11.9  &9.8  &10.4 &14.2  &8.9  &10.3  & ~ \\    %
            
            BiomedParse \cite{biomed-parse} &14.9  &12.2  &9.0  &14.6  &11.7  &9.4  &14.4  &11.5  &9.5  & ~ \\

            IMIS-Net \cite{imis_model} &12.3  &9.6  &11.7  &12.3  &9.6  &16.1  &12.3 &9.8  &16.9  & ~ \\
            
            \specialrule{0em}{0pt}{1pt}
            \hline
            \specialrule{0em}{0pt}{1pt}
            
            DeepSeek-Janus-Pro \cite{deepseek-janus} + BiomedParse \cite{biomed-parse} & 13.9 & 11.4 & 7.6 & 15.4 & 12.5 & 10.3 & 14.2 & 11.5 & 8.5 & \multirow{6}*{\XSolidBrush/\XSolidBrush} \\

            LLaVA-Med \cite{llava-med} + BiomedParse \cite{biomed-parse} & 16.8 & 13.8 & 9.5 & 16.9 & 13.7 & 11.3 & 17.5 & 14.2 & 11.4 & ~ \\
            
            GPT-4o \cite{gpt-4} + BiomedParse \cite{biomed-parse} & 16.7 & 13.8 & 9.3 & 16.1 & 13.1 & 9.8 & 16.9 & 13.8 & 11.1 & ~ \\

            DeepSeek-Janus-Pro \cite{deepseek-janus} + IMIS-Net \cite{imis_model} & 25.2 & 20.7 & 25.6 & 22.8 & 18.7 & 26.5 & 25.5 & 21.3 & 29.1 & ~ \\

            LLaVA-Med \cite{llava-med} + IMIS-Net \cite{imis_model} & 26.1 & 21.6 & 23.6 & 25.8 & 21.3 & 25.2 & 34.0 & 28.9 & 38.3 & ~\\
            
            GPT-4o \cite{gpt-4} + IMIS-Net \cite{imis_model} & 32.4 & 27.0 & 34.5 & 31.8 & 26.7 & 32.7 & 29.7 & 24.8 & 26.0 & ~\\
            
            \specialrule{0em}{0pt}{1pt}
            \hline
            \specialrule{0em}{0pt}{1pt}
            
            LISA-7B \cite{lisa} & 31.6 & 21.2 & 16.2 & 31.1 & 19.9 & 16.0 & 25.4 & 15.5 & 12.2 & \multirow{2}*{\Checkmark/\XSolidBrush} \\
            
            LISA-13B  \cite{lisa} & 35.1 & 23.9 & 19.1 & 35.3 & 22.9 & 18.0 & 28.0 & 17.3 & 13.1 & ~ \\

            \specialrule{0em}{0pt}{1pt}
            \hline
            \specialrule{0em}{0pt}{1pt}
            
            MediSee + READ \cite{lisa-READ} & 55.4 & 45.2 & 54.9 & 54.9 & 44.7 & 60.5 & 54.8 & 44.6 & 60.5 &   \\

            \rowcolor{aliceblue!60}
            MediSee & 56.7 & 46.6 & 56.4 & 57.0 & 46.9 & 61.4 & 57.4 & 47.2 & 61.4 & \Checkmark/\Checkmark\\

            \rowcolor{aliceblue!60}
            MediSee (ft) & \textbf{59.4} & \textbf{49.2} & \textbf{60.3} & \textbf{59.3} & \textbf{49.0} & \textbf{63.3} & \textbf{59.4} & \textbf{49.2} & \textbf{63.7} &  \\

            \bottomrule[1.2pt]      
        \end{tabular}
    }
\vspace{-0.4cm}
\end{table*}

\subsection{Adaptive Democratic Candidate Fusion} \label{sec:Adaptive Democratic Candidate Fusion}
Existing multi-modal large language models (MLLMs) \cite{llava} based methods usually prompt the MLLM to generate a special token (\emph{e.g., [SEG]}) for each target object. Then the downstream decoder \cite{segment-anything} directly uses the hidden representation of this token to produce the mask \cite{lisa++,mmr-iclr2025,multi-round-seg,next-chat}.
Although the aforementioned strategy is straightforward and effective, it imposes the entire responsibility of capturing the necessary hidden features (serving as inputs to the downstream decoder) for each target object onto a single special token.
This heavy reliance on a single token presents challenges in consistently maintaining optimal hidden representations, which may negatively impact overall segmentation performance.
Moreover, when interfacing with multiple downstream decoders, implicit competition may occur among them, resulting in potential performance trade-offs.
Furthermore, the self-optimization process of the special token's features is entirely governed by the intrinsic mechanisms of the large language model (LLM), potentially limiting flexibility and interpretability.

To this end, we propose the Adaptive Democratic Candidate Fusion method, which aims to adaptively and democratically select the optimal features from the candidate tokens, achieving optimizing the tailored inputs for each downstream decoder.
Specifically, we first expand the original LLM vocabulary
with multiple candidate tokens, which meets the request
for the segmentation and detection output.
As shown in Figure \ref{fig:model_main}, given a text instruction \(x_{txt}\) along with the input image \(x_{img}\), we feed them into the MLLM LLaVA-Med \(\mathcal{G}_i\), which in turn outputs a text response \(\hat{y}_{txt}\). It can be formulated as
\begin{align}
\begin{aligned}
    \hat{y}_{txt} = & \;  \mathcal{G}_i(x_{img}, x_{txt}).
\end{aligned}
\end{align}
The output \(\hat{y}_{txt}\) includes the several decoded candidate tokens when the model intends to segment and detect a specific medical entity within \(x_{img}\).
We then extract the LLM last-layer embedding  \(h_{c_i}\) corresponding to the candidate tokens.
To ensure a democratic competition among the features within  \(h_{c_i}\), we employ two lightweight routers, denoted as \(\mathcal{R}_s\) and \(\mathcal{R}_d\), to facilitate the adaptive fusion. 
For each medical object referenced in \(x_{txt}\), the two distinct feature representations \(h_{seg}\) and \(h_{det}\), serving as inputs for the downstream decoders, are computed as weighted sums of the candidate features from \(h_{c_i}\), where the weight is calculated by the routers:
\begin{small}
\begin{equation}
\begin{aligned}
h_{seg}=\sum_{k=1}^n \mathcal{R}_{s}(h_{c_1},h_{c_2}, \dots, h_{c_n})_k \cdot h_{c_k},    \\
h_{det}=\sum_{k=1}^n \mathcal{R}_{d}(h_{c_1},h_{c_2}, \dots, h_{c_n})_k \cdot h_{c_k}.
\end{aligned}
\end{equation}
\end{small}

Finally, the \(h_{det}\) is fed into decoder \(\mathcal{G}_b^{dec}\) to produce bounding box \(\hat{\mathbf{B}}\). While the dense visual features \(f\), extracted by the vision backbone \(\mathcal{G}_i^{enc}\) from the input image \(x_{img}\), are fed into decoder \(\mathcal{G}_i^{dec}\) together with \(h_{seg}\) and \(\hat{\mathbf{B}}\) to generate the segmentation mask \(\hat{\mathbf{M}}\).
The process can be formulated as:
\begin{align}
\begin{aligned}
     f = \mathcal{G}_i^{enc}&\;(x_{img}), \quad               
    \hat{\mathbf{B}} = \mathcal{G}_b^{dec}(h_{det}), \quad \hat{\mathbf{M}} = \mathcal{G}_i^{dec}(f, h_{seg}, \hat{\mathbf{B}}).
\end{aligned}
\end{align}


\subsection{Overall Optimization} \label{sec:Optimization}
\noindent \myparagraph{Training Parameters.}
In order to preserve the fundamental capabilities of the large language model in \(\mathcal{G}_i\) and save computing resources, we leverage LoRA \cite{lora} to perform efficient fine-tuning on the large language model in \(\mathcal{G}_i\), while completely freezing the vision backbone \( \mathcal{G}_i^{enc}\) and vision tower in \(\mathcal{G}_i\).
The mask decoder \( \mathcal{G}_i^{dec}\) is fully fine-tuned.
Additionally, both the two routers ( \(\mathcal{R}_s\) and \(\mathcal{R}_d\)) and the detection decoder \(\mathcal{G}_b^{dec}\) are also fully trainable.

\noindent \myparagraph{End-to-End Training for MediSee.}
MediSee is trained through three supervisions.
As for text generation, we compute the auto-regressive cross-entropy loss \(\mathcal{L}_{txt}\) between the generated text output \(\hat{y}_{txt}\) and the ground-truth text answer \(y_{txt}\).
For the generation of high-quality segmentation masks, the mask loss \(\mathcal{L}_{mask}\) is calculated between the predicted mask \( \hat{\mathbf{M}}\) and the ground-truth mask \( \mathbf{M}\). And the mask loss \(\mathcal{L}_{mask}\) is a weighted sum of the per-pixel binary cross-entropy \(\mathcal{L}_{bce}\) and the DICE loss \(\mathcal{L}_{dice}\), with the weights determined by \(\lambda_{bce}\) and \(\lambda_{dice}\).
For the generation of high-quality bounding boxes, the bounding box loss \(\mathcal{L}_{bbox}\) is calculated between the predicted bounding box \( \hat{\mathbf{B}}\) and the ground-truth bounding box \( \mathbf{B}\), which is a weighted sum of the L1 Loss \(\mathcal{L}_{L1}\) and the GIoU loss \(\mathcal{L}_{giou}\), with the weights determined by \(\lambda_{L1}\) and \(\lambda_{giou}\).
The overall loss \(\mathcal{L}_{end}\) is formulated as
\begin{equation}
\label{eq:total_loss}
\begin{split}
    &\mathcal{L}_{end} = \mathcal{L}_{txt} + \mathcal{L}_{mask} + \mathcal{L}_{bbox}, \\
    &\mathcal{L}_{mask} = \lambda_{bce}\mathcal{L}_{bce} + \lambda_{dice}\mathcal{L}_{dice}, \\
    &\mathcal{L}_{bbox} = \lambda_{L1}\mathcal{L}_{L1} + \lambda_{giou}\mathcal{L}_{giou}. \\
\end{split}
\end{equation}
\noindent \myparagraph{Fine-tuning on MLMR-SD.}
To further explore the upper bound of the proposed MediSee on MedSD task, we additionally fine-tune the model on the proposed MLMR-SD.
As shown in Figure \ref{fig:model_main}, the visual-semantic similarity map corresponding to simple prompts without reasoning is more effective in focusing on the target area.
Therefore, we introduce an additional similarity loss during the fine-tuning phase to further improve the model's comprehensive understanding ability.
Specifically, we compute the similarity matrix between \(\hat{h}_{img}\) and \(\hat{h}_{seg}\) corresponding to the non-inference prompts (\emph{e.g., lung}) as an additional supervisory signal.
Then we adopt loss between $Sim(\hat{h}_{img},\hat{h}_{seg})$ and $Sim({h}_{img},{h}_{seg})$ as \(\mathcal{L}_{sim}\), which is a weighted sum of the JS-Loss \(\mathcal{L}_{js}\) and the MSE loss \(\mathcal{L}_{mse}\), with the weights determined by \(\lambda_{js}\) and \(\lambda_{mse}\):
\begin{equation}
\label{eq:total_loss}
\begin{split}
    &\mathcal{L}_{ft} = \mathcal{L}_{end} + \mathcal{L}_{sim}, \\
    &\mathcal{L}_{sim} = \lambda_{js}\mathcal{L}_{js} + \lambda_{mse}\mathcal{L}_{mse}. \\
\end{split}
\end{equation}

\section{Experiments}

\subsection{Experimental Setting}
\noindent \myparagraph{Network Architecture.}
As mentioned in Sec. \ref{sec:model_architecture}, we adopt LLaVA-Med-v1.5-Mistral-7b \cite{llava-med} as our multi-modal large language foundation model and MedSAM \cite{medsam_model} as our vision backbone.
The projection of \(\mathcal{G}_b^{dec}\) is an MLP with channels of [4, 512, 512, 4096].
The projections of \(\mathcal{R}_s\) and \(\mathcal{R}_d\) are MLPs with channels of [2, 1024, 8192].
In our experiment, we set the parameter \(n\) in the \(h_{c_n}\) to 2, and expand the original LLM vocabulary with two additional candidate tokens.

\noindent \myparagraph{Implementation Details.}
For training MediSee, we utilize 4 NVIDIA A6000 GPUs with 48GB memory, running for approximately 2.5 days. The batch size per device is set to 15.
We adopt the DeepSpeed \cite{deepspeed} engine to enhance the efficiency of our training process. For optimization, we employ the AdamW \cite{adamw} optimizer, with a learning rate of 0.0003. 
The learning rate scheduling is managed by the WarmupDecayLR scheduler, with the warm-up phase configured to span 100 iterations.
The LoRA rank is all set to 8.
The weights for the BCE loss \(\lambda_{bce}\) and the Dice loss \(\lambda_{dice}\) are set to 2.0 and 0.5, respectively. Similarly, the weights for the L1 loss \(\lambda_{L1}\)and the GIoU loss \(\lambda_{giou}\) are both set to 1.0. 
During the additional fine-tuning, the weights for the  JS loss \(\lambda_{js}\) and the MSE loss \(\lambda_{mse}\) are set to 2.0 and 1.0, respectively.

\noindent \myparagraph{Training Datasets.}
We adopt a mixed training dataset, incorporating both the traditional medical semantic segmentation dataset SA-Med2D-20M \cite{sam_med_2d_20m} and our MLMR-SD dataset.
We utilize a subset of SA-Med2D-20M for training and employ a series of carefully designed templates with the original labels within the dataset to generate query inputs.
The sampling ratio of the two datasets is set to \(7:3\).

\begin{table*}[t]
    \centering
    \caption{Comparison of medical reasoning detection performance between MediSee and prior related methods. \(\mathcal{E}\) and \(\mathcal{G}\) denote the explanation output and the generation type of bounding box, respectively. ``mask2box'': manually deriving bounding boxes from masks; ``decoder'': bounding boxes directly output by the decoder.}
    \vspace{-0.2cm}
    \label{table:reason_box}   
    \tabcolsep=0.42cm
    {
        \begin{tabular}{ l | c c | c c | c c | c  }
            \toprule[1.2pt]
            
            \multirow{3}*{\textbf{Methods}} & \multicolumn{2}{c|}{\textbf{val}} & \multicolumn{4}{c|}{\textbf{test}}  & \multirow{3}*{\(\mathcal{E}\)/\(\mathcal{G}\)}     \\

            \specialrule{0em}{0pt}{1pt}
            \cline{2-7}
            \specialrule{0em}{0pt}{1pt}
            
            
            ~ & \multicolumn{2}{c|}{\textbf{overall}} & \multicolumn{2}{c|}{\textbf{short query}} & \multicolumn{2}{c|}{\textbf{long query}} & ~ \\

            \specialrule{0em}{0pt}{1pt}
            \cline{2-7}
            \specialrule{0em}{0pt}{1pt}
            
            ~ & \textbf{IoU} & \textbf{Acc} & \textbf{IoU} & \textbf{Acc} & \textbf{IoU} & \textbf{Acc}  & ~  \\ 
            
            \specialrule{0em}{0pt}{1pt}
            \hline
            \hline
            \specialrule{0em}{0pt}{1pt}

            Grounding DINO \cite{grounding-dino} + SAM-Med2D \cite{sam-med-2d-model} &12.1  &10.8  &9.7  &6.7  &14.1  &12.1    & \multirow{4}*{\XSolidBrush/mask2box}\\

            Grounding DINO \cite{grounding-dino} + MedSAM \cite{medsam_model} &10.0  &7.1  &11.5 &7.4  &15.7 &12.5  & ~ \\    %
            
            BiomedParse \cite{biomed-parse} &13.5  &14.1  &13.1  &13.3  &12.9  &12.9   & ~ \\

            IMIS-Net \cite{imis_model} &11.7  &10.9  &12.1  &11.0  &12.3  &10.9   & ~ \\
            
            \specialrule{0em}{0pt}{1pt}
            \hline
            \specialrule{0em}{0pt}{1pt}
            
            DeepSeek-Janus-Pro \cite{deepseek-janus} + BiomedParse \cite{biomed-parse} &12.5  &12.9  &13.8  &14.0  &12.7  &12.9   & \multirow{6}*{\XSolidBrush/mask2box} \\

            LLaVA-Med \cite{llava-med} + BiomedParse \cite{biomed-parse} &14.8  &15.5  &15.1  &15.5  &15.6  &16.4  & ~ \\
            
            GPT-4o \cite{gpt-4} + BiomedParse \cite{biomed-parse} &15.1  &15.8  &14.7  &15.2  &15.4  &16.0  & ~ \\

            DeepSeek-Janus-Pro \cite{deepseek-janus} + IMIS-Net \cite{imis_model} &23.5  &24.9  &21.4  &21.7  &24.1  &24.7  & ~ \\

            LLaVA-Med \cite{llava-med} + IMIS-Net \cite{imis_model} &24.2  &25.4  &23.9  &24.6  &31.7 &33.2  & ~\\
            
            GPT-4o \cite{gpt-4} + IMIS-Net \cite{imis_model} &29.9  &32.5  &29.8  &31.7  &27.7  &29.5  & ~\\
            
            \specialrule{0em}{0pt}{1pt}
            \hline
            \specialrule{0em}{0pt}{1pt}
            
            Uni-Med \cite{uni-med} & 21.1 & 13.2 & 21.5 & 10.7 & 22.6 & 11.8  & \multirow{2}*{\XSolidBrush/decoder} \\
            
            LLaVA-Med \cite{llava-med} + Uni-Med \cite{uni-med} & 27.9 & 17.1 & 27.8 & 18.9 & 27.7 & 19.3 & ~ \\

            \specialrule{0em}{0pt}{1pt}
            \hline
            \specialrule{0em}{0pt}{1pt}
            
            MediSee + READ \cite{lisa-READ} & 34.6 & 31.0 & 35.1 & 32.6 & 35.0 & 31.5 &  \\
            \rowcolor{aliceblue!60}
            MediSee & 35.7 & 32.4 & 36.2 & 33.5 & 36.7 & 34.5 & \Checkmark/decoder\\

            \rowcolor{aliceblue!60}
            MediSee (ft) & \textbf{37.8} & \textbf{36.7} & \textbf{38.3} & \textbf{37.0} & \textbf{38.5} & \textbf{36.3}  &  \\
            
            \bottomrule[1.2pt]      
        \end{tabular}
    }
\vspace{-0.4cm}
\end{table*}

\subsection{Medical Reasoning Segmentation Results} 
\label{sec:med-reasin-seg-exp}
The medical reasoning segmentation results are presented in Table \ref{table:reason_seg}. 
Due to the fact that related works based on SAM \emph{e.g.,} MedSAM \cite{medsam_model} and SAM-Med2D \cite{sam-med-2d-model} are predominantly designed for bounding box inputs, we consider utilizing Grounding DINO \cite{grounding-dino} to add a text understanding interface to these methods.
Specifically, we first use Grounding DINO to generate a corresponding bounding box range based on the query then input it as instructions to MedSAM and SAM-Med2D to obtain the segmentation masks.
As shown in Table \ref{table:reason_seg}, existing works fail to address this task, as they are unable to truly comprehend queries that involve complex logical reasoning and identify the corresponding masks in medical images. 
In contrast, the proposed MediSee effectively integrates hidden information from different candidate tokens to capture the target semantic features corresponding to the medical images, effectively achieving this capability and demonstrates an average improvement of about 20\% across various metrics compared to other methods.

Notably, we also make a combined approach (MLLM + medical segmentation model) as a competitor.
Specifically, we borrow the powerful text-image understanding capabilities of MLLM \cite{deepseek-janus,llava-med,gpt-4} to preliminarily process the complex queries, and then feed the generated textual reasoning outputs into the medical segmentation model.
The results presented in Table \ref{table:reason_seg} demonstrate that, despite being equipped with powerful MLLMs, existing medical segmentation models still underperform compared to MediSee.
We owe this success to the fact that our model is trained end-to-end on a large-scale dataset of medical image-text pairs, which enables a more seamless and coherent feature flow across different modalities compared to combined models.
Additionally, we report LISA \cite{lisa}, a representative work in natural image reasoning segmentation, in our comparative analysis. The results demonstrate that, although LISA excels in natural image reasoning segmentation, it struggles in medical contexts.

Furthermore, it is noteworthy that existing medical segmentation models rarely incorporate both an independent mask decoder and a bounding box decoder, and they often fail to provide necessary explanations for complex queries. 
In contrast, our model simultaneously integrates these capabilities.

    

        
        
        
        

\subsection{Medical Reasoning Detection Results}
Similar to Section \ref{sec:med-reasin-seg-exp}, we present the medical reasoning detection results in Table \ref{table:reason_box}.
Most existing medical segmentation models do not have an independent bounding box decoder, which makes them not directly comparable in detection tasks. 
We circumvent this by manually extracting the minimum bounding box that covers the medical object mask from the segmentation outputs for comparison.
The results presented in Table \ref{table:reason_box} demonstrate that MediSee significantly outperforms other methods. 
We attribute this performance improvement to the following key factors: 
1) Our model is capable of understanding implicit queries that involve medical image detection requirements.
2) Our model incorporates a specialized medical detection decoder that is independent of the mask decoder, ensuring more accurate detection results.
3) The Adaptive Democratic Candidate Fusion method we employ not only enables the model to effectively capture higher-quality features to downstream decoders, but also facilitates the simultaneous balancing of multiple tasks, including both segmentation and detection.

Moreover, we compare MediSee with other related methods \cite{uni-med} equipped with the ability of medical referring expression comprehension in Table \ref{table:reason_box}. 
The results also indicate that MediSee exhibits superior medical reasoning detection capabilities.

\subsection{Vanilla Medical Referring Perception}
To prove that MediSee is also competent in vanilla referring medical segmentation and detection, we compare it with existing methods on SA-Med2D-20M \cite{sam_med_2d_20m} benchmark.
As shown in Table \ref{table:traditional_seg_box}, our approach remains highly competitive in traditional medical image perception task.
Moreover, we additionally present the segmentation results of IMIS-Net \cite{imis_model} and MedSAM \cite{medsam_model} under traditional bounding box guidance to give more insights.

\begin{table}[t]
    \footnotesize
    \centering
    \caption{Vanilla medical referring perception results. } 
    \vspace{-0.2cm}
    \label{table:traditional_seg_box}   
    \tabcolsep=0.153cm
    {
        \begin{tabular}{ l | c c c | c c  }
            \toprule[1.2pt]
            
            \multirow{2}*{\textbf{Methods}} & \multicolumn{3}{c|}{\textbf{Seg.}} & \multicolumn{2}{c}{\textbf{Det.}}      \\ 
            

            \specialrule{0em}{0pt}{1pt}
            \cline{2-6}
            \specialrule{0em}{0pt}{1pt}
            
            ~ & \textbf{Dice} & \textbf{gIoU} & \textbf{cIoU} & \textbf{IoU} & \textbf{Acc}  \\ 
            
            \specialrule{0em}{0pt}{1pt}
            \hline
            \hline
            \specialrule{0em}{0pt}{1pt}

             \multicolumn{6}{l}{\emph{{\textbf{Medical Interaction Seg. (bbox input)  }}}} \\
             
            \largemodel MedSAM \cite{medsam_model} &\largemodel 77.7 &\largemodel 67.1 &\largemodel 81.0 &\largemodel - &\largemodel - \\

            \largemodel IMIS-Net \cite{imis_model} &\largemodel 77.0 &\largemodel 67.5 &\largemodel 87.2 &\largemodel - &\largemodel -  \\

            \specialrule{0em}{0pt}{1pt}
            \hline
            \specialrule{0em}{0pt}{1pt}
            
             \multicolumn{6}{l}{\emph{{\textbf{Medical Referring Seg. \& Det.  }}}} \\
            Grounding DINO \cite{grounding-dino} + SAM-Med2D \cite{sam-med-2d-model} & 18.4 & 12.5 & 10.9 & 14.8 & 13.3  \\

            Grounding DINO \cite{grounding-dino} + MedSAM \cite{medsam_model} & 18.5 & 11.0 & 13.5 & 17.3 & 15.3  \\    %
            
            BiomedParse \cite{biomed-parse} & 24.9 & 20.1 & 14.5 & 26.0 & 23.7 \\

            Uni-Med \cite{uni-med} &- &- &-  & 31.4 & 24.6  \\

            IMIS-Net \cite{imis_model} &41.3  & 35.5 & 65.8 & 39.3 & 39.6  \\
     
            
            MediSee + READ \cite{lisa-READ} & 60.3 & 49.2 & 69.4 & 43.3 & 47.5  \\
            

            \rowcolor{aliceblue!60}
            MediSee & \textbf{61.2} & \textbf{50.4} & \textbf{70.8} & \textbf{45.6} & \textbf{51.0}   \\
            
            \bottomrule[1.2pt]      
        \end{tabular}
    }
\end{table}

\begin{table}[t]
    \footnotesize
    \centering
    \caption{Ablation study on our Adaptive Democratic Candidate Fusion method.} 
    \vspace{-0.2cm}
    \label{table:moe_ablation}   
    \tabcolsep=0.13cm
    {
        \begin{tabular}{ c | c | c | c c c c   }
            \toprule[1.2pt]
            
            \textbf{(\(n\) in \(h_{c_n}\))} & \textbf{Fusion} & \textbf{hidden size}  & \textbf{gIoU} & \textbf{cIoU} & \textbf{IoU} & \textbf{Acc}       \\ 
            
            \specialrule{0em}{0pt}{1pt}
            \hline
            \specialrule{0em}{0pt}{1pt}

            1 & hard &- & 49.8/45.4 & 69.7/55.1 & 43.6/33.4 &47.5/30.9    \\
            
            \specialrule{0em}{0pt}{1pt}
            \hline
            \specialrule{0em}{0pt}{1pt}
            
             \multirow{4}*{2} & hard &- & 49.3/46.1 & 70.2/56.2 &44.1/34.9 &50.5/32.2  \\

            ~ &  & 512 & 49.5/44.5 & 68.3/55.2 &42.4/34.2 &45.5/30.8\\

           &\cellcolor{aliceblue!60}soft  &\cellcolor{aliceblue!60}1024 
           &\cellcolor{aliceblue!60}50.4/46.6  &\cellcolor{aliceblue!60}70.8/56.4    &\cellcolor{aliceblue!60}45.6/35.7   &\cellcolor{aliceblue!60}51.0/32.4 \\
            
           &  & 2048 & 50.0/46.3 & 70.5/56.0 & 43.8/35.4 &48.5/32.0 \\

            \specialrule{0em}{0pt}{1pt}
            \hline
            \specialrule{0em}{0pt}{1pt}

             3 & soft &1024 & 45.1/35.7 & 64.1/46.1 & 40.6/31.0 &40.3/25.9  \\
            
            \bottomrule[1.2pt]      
        \end{tabular}
    }
\vspace{-0.4cm}
\end{table}

\begin{figure*}
    \centering
    \includegraphics[width=1\linewidth]{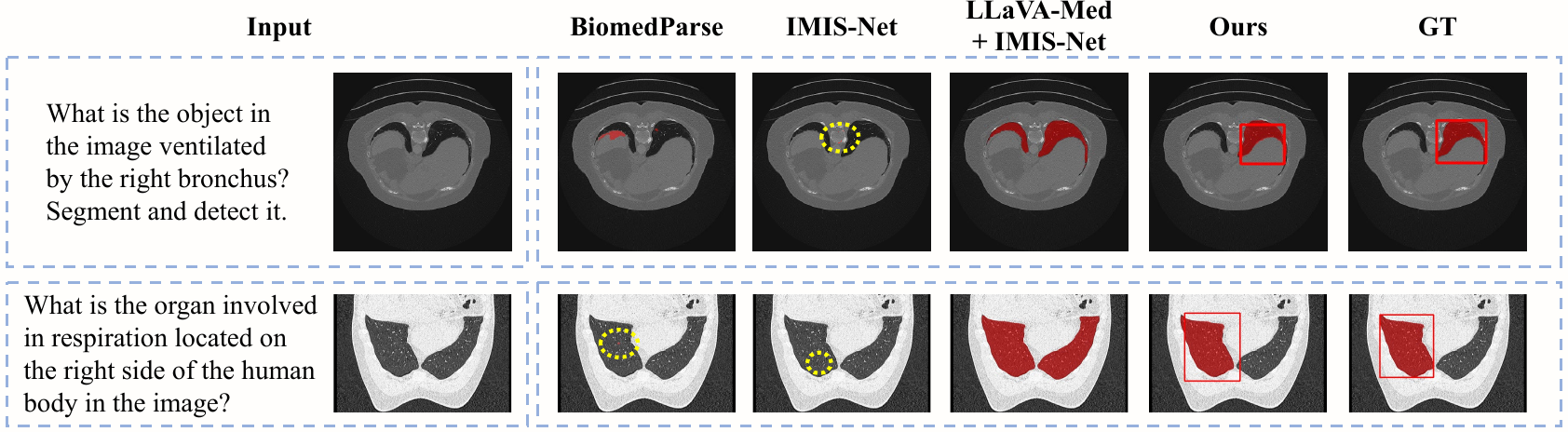}
    \vspace{-2.1em}
    \caption{Visualizations of different methods. Some mask areas with tiny response are marked by yellow dashed lines.}
    \vspace{-0.5em}
    \label{fig:visual_perform_exp}
    \Description{}
\end{figure*}

\subsection{Ablation Study}
We conduct a comprehensive ablation study to assess the contribution of each component. The results are reported on the MLMR-SD validation set and SA-Med2D-20M test set. 
The former of the slashes ``/'' in Tables ~\ref{table:moe_ablation}, ~\ref{table:box_text_point_2samed}, and ~\ref{table:mllm_backbone_and_seg_and_simloss} represents the results based on the traditional medical referring perception, while the latter represents the results based on MedSD setting.

\noindent \myparagraph{Ablation on Adaptive Democratic Candidate Fusion.}
We analyze the effectiveness of our Adaptive Democratic Candidate Fusion method from two perspectives: the number of candidate tokens (\(n\) in the \(h_{c_n}\)) and the way of feature fusion (soft or hard).
In Table \ref{table:moe_ablation}, we notice that when \(n=2\) and the hidden layer dimension of \(\mathcal{R}_s\) and \(\mathcal{R}_d\) is 1024, MediSee performs the best.
This is potentially because our architecture incorporates two downstream decoders, and the 1024-dimensional hidden layer effectively captures and integrates the feature information from the candidate tokens.

\noindent \myparagraph{Input Choices of MedSAM.}
We experimented with providing MedSAM different types of guidance information. 
Notably, the bounding box information is the output of \(\mathcal{G}_b^{dec}\), while the point information is obtained using the READ \cite{lisa-READ} method.
The results presented in Table \ref{table:box_text_point_2samed} indicate that the model performs optimally when both text embedding and bounding box information are provided simultaneously. 
This suggests that more guidance information does not necessarily for MedSAM lead to better performance.

Moreover, we observe that the version of \((\mathcal{T} + \mathcal{B})\) also exceeds version \((\text{only }\mathcal{T})\) by approximately 2.6\% on detection metrics in average.
This improvement may arise from the fact that during training, inputting MedSAM with bounding box information not only serves as additional guidance but also enables the backward optimization of the detection decoder's output.

\noindent \myparagraph{Choices of Segmentation Head.}
In Table \ref{table:mllm_backbone_and_seg_and_simloss}, we investigate the performance impact of different models as segmentation head backbones. 
The results show that the MedSAM-based version observes the best performance, likely due to the advantages gained from its pretraining on a large scale of medical images.
Furthermore, we also find that training MedSAM with LoRA does not yield performance as effective as full fine-tuning.

\noindent \myparagraph{Choices of MLLM Backbone.}
Table ~\ref{table:mllm_backbone_and_seg_and_simloss} presents the impact of different MLLMs as the vision-language foundational models. The results show that LLaVA-Med outperforms better, which can be attributed to its inherent ability to align multi-modal features in the medical context. 
Moreover, we observe that freezing the parameters of LLAVA-Med leads to a decline in the model's performance.

\noindent \myparagraph{Impact of Similarity Loss.}
When further fine-tuning the model on the MLMR-SD dataset, we introduce a similarity loss as an additional supervisory signal. The results in Table ~\ref{table:mllm_backbone_and_seg_and_simloss} demonstrate that, with the incorporation of similarity loss, the performance improves by 1.8\% in terms of segmentation metrics and 1.1\% in terms of detection metrics, thereby validating its effectiveness.

\subsection{Qualitative Analysis}
We visualize some cases of the MLMR-SD validation split in Figure \ref{fig:visual_perform_exp} to give some insights.
As shown in Figure \ref{fig:visual_perform_exp}, existing methods are largely incapable of understanding implicit queries regarding medical images. Even with the assistance of MLLMs, the quality of the generated masks is lower than our MediSee's result.
Notably, MediSee is capable of generating high-quality masks and bounding boxes while simultaneously providing text explanations relevant to complex queries.
More dialogue demos about MediSee can be found in the \textit{supplementary materials}.

\begin{table}[t]
    \footnotesize
    \centering
    \caption{Ablation study on the input of MedSAM. \(\mathcal{T}\): text embedding; \(\mathcal{B}\): bounding box; \(\mathcal{P}\): point.}
    \vspace{-0.2cm}
    \label{table:box_text_point_2samed}
    \tabcolsep=0.3cm
    \begin{tabular}{c c c | c c c c}
        \toprule[1.2pt]
         \(\mathcal{T}\) & \(\mathcal{B}\) & \(\mathcal{P}\) &\textbf{ gIoU} & \textbf{cIoU} &\textbf{ IoU} &\textbf{ Acc} \\
        \midrule
        \Checkmark  & & & 49.3/46.5 & 70.0/57.1  & 44.3/35.4 & 46.8/31.5 \\

        \rowcolor{aliceblue!60}
        \Checkmark &  \Checkmark & & 50.4/46.6 & 70.8/56.4 & 45.6/35.7 &51.0/32.4  \\
         \Checkmark & \Checkmark & \Checkmark & 49.2/45.2 & 69.4/54.9  & 43.3/34.6 & 47.5/31.0 \\
        \bottomrule[1.2pt]
    \end{tabular}
\end{table}

\begin{table}[t]
    \footnotesize
    \centering
    \caption{Ablation study on the choices of vision backbone and MLLM backbone, and the impact of \( \mathcal{L}_{sim}\) during additional fine-tuning.}
    \vspace{-0.2cm}
    \label{table:mllm_backbone_and_seg_and_simloss}
    \tabcolsep=0.28cm
    \begin{tabular}{l | c c c c}
        \toprule[1.2pt]

        \textbf{Setting} & \textbf{gIoU} & \textbf{cIoU}  & \textbf{IoU} & \textbf{Acc} \\
        \midrule
        \midrule

         \multicolumn{5}{l}{\emph{{\textbf{Seg. Head Backbone}}}} \\
         SAM & 36.6/29.1 & 57.0/35.9   & 19.3/16.1 & 6.5/6.9\\
        
         SAM-Med2D  & 46.5/41.1 & 66.5/51.5  &44.9/34.8  &49.0/32.1 \\
        
         MedSAM (LoRA) & 46.3/40.8 & 66.2/52.3 &44.9/35.5  &50.0/32.5 \\

          \rowcolor{aliceblue!60}
         MedSAM  & 50.4/46.6 & 70.8/56.4 & 45.6/35.7 &51.0/32.4 \\

        \specialrule{0em}{0pt}{1pt}
        \hline
        \specialrule{0em}{0pt}{1pt}
        
         \multicolumn{5}{l}{\emph{{\textbf{MLLM Backbone}}}} \\
        LLaVA-v1.5 (LoRA) & 49.5/43.2 & 68.9/52.6   & 43.7/32.8 & 49.0/28.9\\
        
         LLaVA-Med (frozen) & 43.6/37.3 & 63.9/48.5  &27.3/22.7   &18.0/14.2 \\
         
        \rowcolor{aliceblue!60}
         LLaVA-Med (LoRA) & 50.4/46.6 & 70.8/56.4 & 45.6/35.7 &51.0/32.4 \\

        \specialrule{0em}{0pt}{1pt}
        \hline
        \specialrule{0em}{0pt}{1pt}
         \multicolumn{5}{l}{\emph{{\textbf{Fine-tuning Loss}}}} \\
         w/o \( \mathcal{L}_{sim}\)  & 48.3 & 57.6   &37.2   &35.1 \\

        \rowcolor{aliceblue!60}
        w/ \( \mathcal{L}_{sim}\) & 49.2 & 60.3  &37.8  &36.7  \\
        \bottomrule[1.2pt]
    \end{tabular}
\vspace{-0.2cm}
\end{table}

\section{Conclusion}

This work introduces a novel medical vision task, Medical Reasoning Segmentation and Detection (MedSD), designed to interpret implicit queries about medical images and generate the corresponding segmentation mask and bounding box. 
We also propose a medical reasoning image perception dataset as the research foundation, featuring a large collection of medical entity targets and their corresponding reasoning-based question-answer pairs. 
Additionally, we present an effective baseline model for the MedSD task termed MediSee, which is capable of reasoning complex queries related to medical images.
We hope this work can provide some insights for bringing medical segmentation models into everyday life.

\clearpage

\input{Main.bbl}

\bibliographystyle{IEEEtran}

\end{document}